\documentclass[letterpaper, 10 pt, conference]{ieeeconf}  

\IEEEoverridecommandlockouts                              

\makeatletter
\let\NAT@parse\undefined
\makeatother
\usepackage[hidelinks]{hyperref}  
\usepackage{xurl}
\usepackage{microtype}
\usepackage{gensymb}
\usepackage{amsmath,amssymb,amsfonts}
\usepackage{bm}
\usepackage{algorithm}
\usepackage[noend]{algpseudocode}

\usepackage{tabularx}
\usepackage{multirow}
\usepackage{graphicx}
\makeatletter
\let\MYcaption\@makecaption
\makeatother
\usepackage[font=footnotesize]{subcaption}
\makeatletter
\let\@makecaption\MYcaption
\makeatother
\usepackage{textcomp}
\usepackage{xcolor}

\makeatletter
\newcommand{\multiline}[1]{%
  \begin{tabularx}{\dimexpr\linewidth-\ALG@thistlm}[t]{@{}X@{}}
    #1
  \end{tabularx}
}
\makeatother

\title{\LARGE \bf
OA-Bug: An Olfactory-Auditory Augmented Bug Algorithm\\
for Swarm Robots in a Denied Environment
}

\author{Siqi Tan, Xiaoya Zhang, Jingyao Li, Ruitao Jing, Mufan Zhao, Yang Liu, and Quan Quan* 
\thanks{This work was done at Beihang University and was supported by the National Natural Science Foundation of China under Grant 61973015.}
\thanks{*Corresponding author. Email: \href{mailto:qq_buaa@buaa.edu.cn}{\texttt{qq\_buaa@buaa.edu.cn}}.}
}

\begin{document}
\maketitle
\thispagestyle{empty}
\pagestyle{empty}

\begin{abstract}

Searching in a denied environment is challenging for swarm robots as no assistance from GNSS, mapping, data sharing, and central processing is allowed. However, using olfactory and auditory signals to cooperate like animals could be an important way to improve the collaboration of swarm robots. In this paper, an Olfactory-Auditory augmented Bug algorithm (OA-Bug) is proposed for a swarm of autonomous robots to explore a denied environment. A simulation environment is built to measure the performance of OA-Bug. The coverage of the search task can reach 96.93\% using OA-Bug, significantly outperforming a similar algorithm, SGBA~\cite{c1}. Furthermore, experiments are conducted on real swarm robots to prove the validity of OA-Bug. Results show that OA-Bug can improve the performance of swarm robots in a denied environment.

Video: \url{https://youtu.be/vj9cRiSm9eM}.

\end{abstract}

\section{Introduction}

Robotic search and rescue (RSAR) holds great promise compared to human search and rescue due to the efficiency of a multi-robot system (MRS), as well as risk and cost reduction related to human resecurs. The MRS is formed by swarms of autonomous robots with positioning equipment and other essential sensors, which are well-trained to perform their respective functions. This guarantees a faster response~\cite{c2,c3} and the obtainability of real-time mapping and the monitoring of the accident scene~\cite{c4,c5} or even proving of support in hazardous environments~\cite{c6,c8}.

\textls[-1]{Some researchers have provided feasible solutions for RSAR to reduce the time cost while improving the search coverage~\cite{c1,c9,c10,c11,c12}. Others have developed strategies to enhance the robustness of MRS~\cite{c13,c14}. However, these solutions are based on external resources or prior knowledge of the rescue scene, enabling robots to pre-plan their paths. During a rescue, the robots only need to follow a calculated route with the help of the Global Navigation Satellite System (GNSS) or the pre-built map without decisions being made based on the local environment. However, the GNSS signal may be unavailable for various reasons, and the rescue scene may be unfamiliar to humans and robots. In addition, some algorithms require central processing or data sharing among robots, which needs communication over the network and may also be impractical in poor communication conditions.}

\textls[-1]{We define a “denied environment” as an unknown environment where global positioning is unavailable and communication is constrained. Prior to our research, a graph-based model was proposed for multi-robots to cover a denied environment~\cite{c15}, and robots were designed to utilize RFID tags as coordination points to memorize and update their pose~\cite{c16}. However, these graph-based approaches need internal memory and can be computationally expensive and inapplicable in large complex environments~\cite{c17}. Later, a robot that can repeatedly leave pen-marked trails in the terrain to cover a closed terrain was built~\cite{c18}. However, pen-marked tracks can be hard to detect in dirty or dark environments. MAW~\cite{c19}, StiCo~\cite{c17} and S-MASA~\cite{c20} are stigmergic approaches without the graph model. However, these approaches have not been implemented in hardware. MAW~\cite{c19} needs to find a new direction with the lowest level of pheromone in each move, which may be too time-consuming to be practical in real-world implementations. StiCo~\cite{c17} and S-MASA~\cite{c20} are not validated in complex sites with structures like nested rooms, and the robots may seek to fully cover the space in such rooms, which is unnecessary and may result in getting stuck. Besides, the above methods fail to consider how to choose initial directions when robots depart together, and ignore other robots’ relative positions when robots choose new directions.}

\begin{figure}[t]
  \vspace{6pt}
  \centerline{\includegraphics[width=0.49\textwidth]{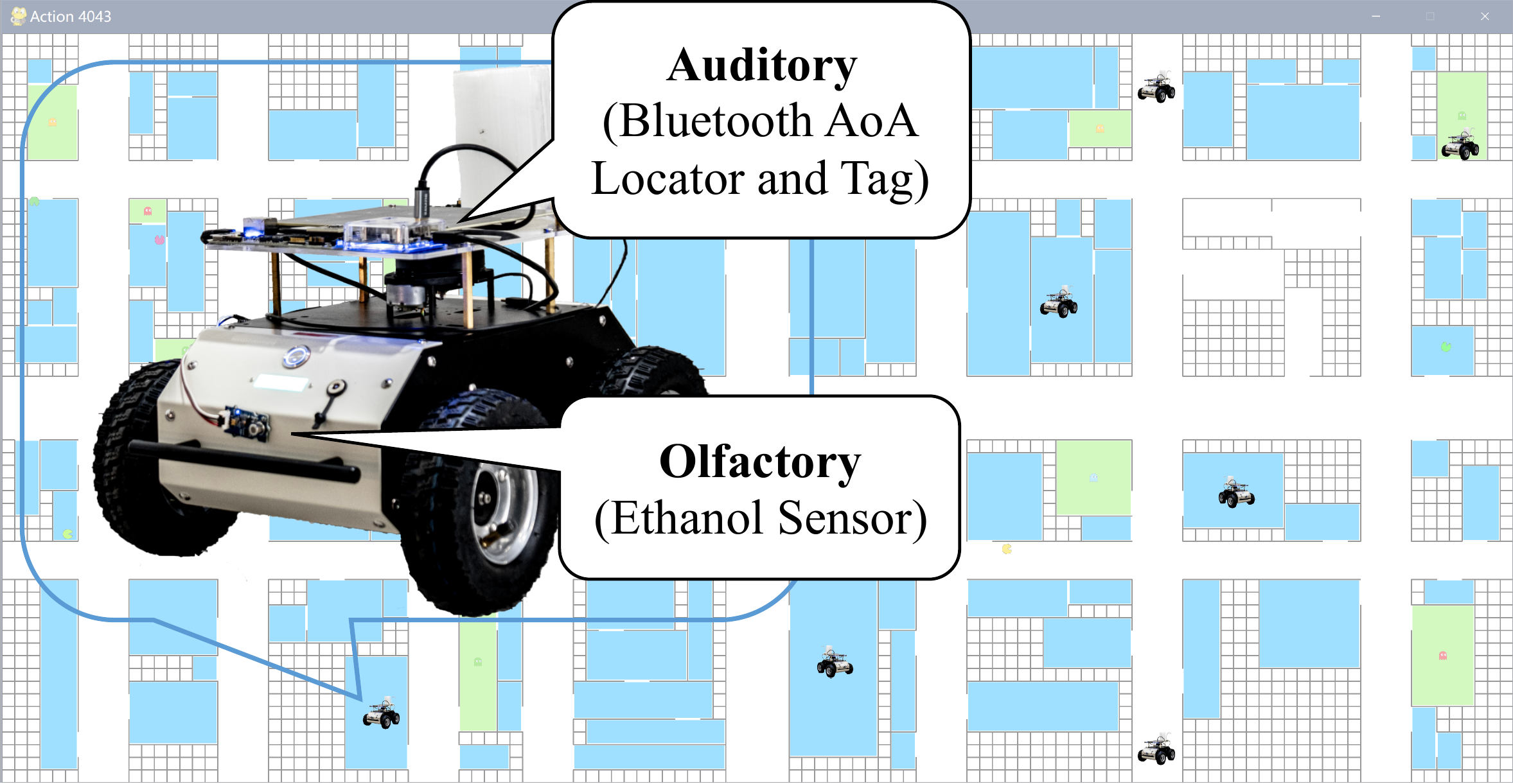}}
  \caption{Swarm Robots with Olfactory and Auditory}
  \label{header}
  \vspace{-15pt}
\end{figure}

\textls[-2]{The “Bug Algorithm” is a traditional algorithm for robotic motion planning and target tracking~\cite{c21}. Only requiring local sense and the direction to the goal, it is a simple method without the need to worry about resource constraints when implemented on hardware. Many variants were proposed to improve its performance~\cite{c22}, including vision-based ones that detect and memorize scene feature descriptors using SIFT~\cite{c22.1} and Bag-of-Words~\cite{c22.2}. But vision-based methods require extra memory and may not work in poor lighting conditions where environment textures are unrecognizable. SGBA~\cite{c1} is a solution for a swarm of robots to explore a denied environment. However, a centralized Wi-Fi home beacon is still needed to provide critical information, which may still be inapplicable in some cases due to the lack of external power supply, deployment difficulties, signal interference, and shielding.}

\textls[-2]{In denied environments, humans, as well as animals, usually acquire and transmit effective information through olfactory and auditory. For example, search and rescue team members in a cave can call out to each other to know their relative positions; animals in the forest that are blocked from seeing each other can transmit information through chirping; ants release pheromones to transmit information through chemicals with the “hysteresis effect.” Can olfactory and auditory be combined with the “Bug Algorithm” and implemented on real robots?}

In terms of olfactory, a low-cost gas sensor has been deployed~\cite{c23} on a robot to search for the gas source in a turbulent and diffusive environment. Even circuits or robots have been equipped with insects’ odor-sensing antennae to perform specific tasks~\cite{c24,c25,c26,c27}. Thus, olfactory is feasible for robots. As for auditory, ODAS~\cite{c28,c29} is a powerful system for sound source localization (SSL) and tracking. The system itself is well optimized for accurate, fast and robust SSL performance, but in the first iteration of our deployment, we found that using sound was susceptible to interference and the transmission distance was too short. Fortunately, the new Bluetooth 5.1 Direction Finding standard can provide us with higher precision, a much longer range, and more robust source localization and tracking performance~\cite{c30}, making auditory feasible for swarm robots.

In this paper, we propose an Olfactory-Auditory Augmented Swarm Bug Algorithm (OA-Bug) for a swarm of autonomous robots to explore an unknown and unstructured environment efficiently without the assistance of GNSS, mapping, data sharing, or central processing. Each robot leaves a scent, such as ethanol, on its path to mark that place as visited, which other robots can detect. Once detected, the robot can receive auditory signals from other robots to obtain their relative positions, facilitating its orientation.

\textls[-5]{Compared with other swarm algorithms, our algorithm has higher coverage within a reasonable period of time for searching in a denied environment. It is known that the main causes of death in disasters are delays in rescue efforts and the inability to find victims~\cite{c31}. Thus, our algorithm has a better chance of ensuring the survival of the victims in real-life scenarios.}

\section{Problem Statement and\\Algorithm Formulation}

\subsection{Problem Statement}\label{defs}

Consider a scene consisting of a static site $S=\{w_1,w_2,\\\dots,w_n,a_1,a_2,\dots,a_m,B\}$ which is a flat figure, a swarm of robots $R=\{r_1,r_2,\dots,r_p\}$, and optional static victims $V=\{v_1,v_2,\dots,v_q\}$, where $w_i  (1<i \le n)$ is each wall, $a_i  (1<i \le m)$ is each room area, $B$ is all blank areas that are not supposed to be visited, $r_i  (1<i \le p)$ is the $i$th robot, and $v_i  (1<i \le q \le p)$ is the $i$th victim that does not move. These notations are simply identifiers for convenience.

\begin{figure}[t]
  \vspace{5pt}
  \centerline{\includegraphics[width=0.49\textwidth]{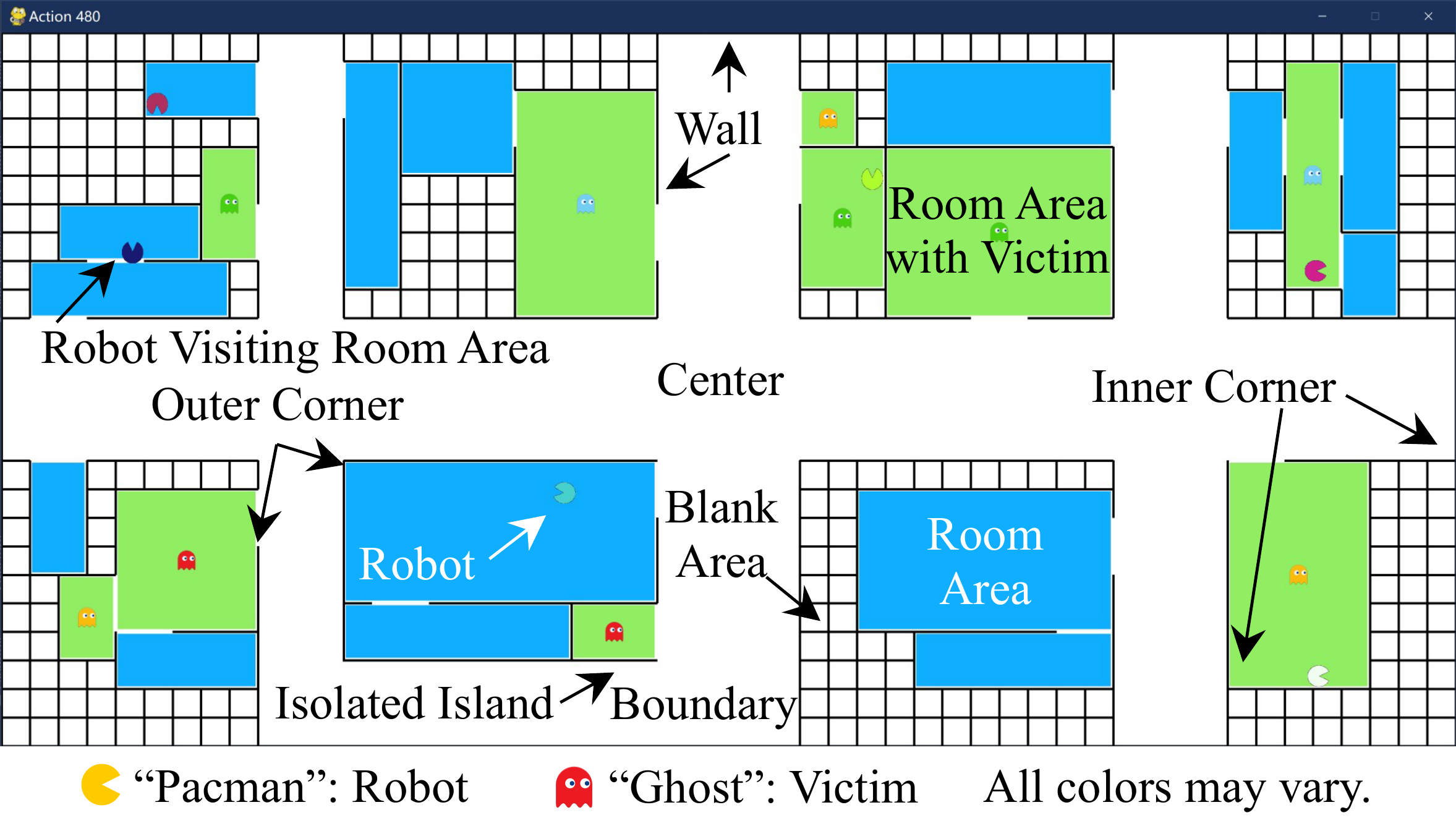}}
  \vspace{-9pt}
  \caption{Site Definitions and Descriptions}
  \label{defs fig}
  \vspace{-17pt}
\end{figure}

The site has two features. One feature is unfamiliarity, meaning that robots have never been there before, and there are no auxiliary devices for positioning and communication on the site. The other feature is closure, indicating the site is surrounded by walls that robots cannot destroy, and there is only one entrance or exit for robot deployment.

\textls[-1]{We define the projection of a “wall” on the site as a continuous curve with two ends called “endpoints.” We define a “room” as a union of walls. The projection of a “room” on the site is a flat figure with at least one interruption, i.e., at least two endpoints, and the total number of endpoints is a multiple of two; a way can be found to connect the endpoints, forming only a minimum enclosed figure. We define this enclosed figure as a “room area.” We define an “isolated island” as a set of rooms that are connected but have no walls leading to the walls surrounding the site. To make it easier and more realistic, we assume walls are straight segments, and we define the smaller angle formed by the intersection of two walls as an “inner corner of the walls” and the larger one as an “outer corner of the walls.” Some definitions are shown in Fig.~\ref{defs fig}.}

All robots are identical, i.e., having the same action space ${A} = \{\text{go straight}, \text{turn}\}$ and state space, as well as limited and equal sensing abilities. Robots need to set off from a designated location in the site (e.g., the “center” or the “boundary”). Their task is “searching,” that is, searching as many unvisited room areas as possible in a period of time to find victims (if any). There is no specific goal for each robot in the search task, during which they cannot memorize or share their exact states. In our setting, if no other sensing abilities are added, the robots can only know whether they are hitting the wall or other robots, whether they are at the outer corner of the walls, and whether they have found victims.

A robot needs to have the following four types of sensors.

\textit{1) Ranging sensor:} It allows robots to know whether they are hitting a wall or other robots, whether they are at an outer corner of walls, and whether they have found victims. It returns the relative position of each detected object named above with respect to the position $\boldsymbol{p}_i$ of the source robot $r_i$.

\textit{2) Olfactory sensor:} It returns the number of times the current location $l_{ci} \in S$ has been visited.

\textit{3) Auditory sensor:} It allows robots to get others’ relative positions. It returns $\boldsymbol{p}_j = (\rho_j, \theta_j) (j \ne i)$ where $\rho_j$ is the distance to the $j$th robot and $\theta_j$ is the azimuth of the $j$th robot with respect to the position $\boldsymbol{p}_i$ of the source robot $r_i$.

\textit{4) Orientation sensor:} It allows a robot to turn towards an azimuth. It returns the current azimuth of the robot.

The Bug Algorithm assumes idealistic sensors that do not involve uncertainties. We therefore also adhere to this assumption in order to avoid implementation details interfering with the algorithm design.

\subsection{Algorithm Formulation}

We follow the principles below to design our algorithm.

\textit{1) Reducing Revisiting:} A place that a robot has visited should not be visited again and again.

\textit{2) Reducing Clustering:} Robots should spread evenly.

\textit{3) Searching as Much as Possible:} If a place has not been visited, the robot should continue to search it rather than change to a new direction.

\begin{figure}[t]
  \vspace{3pt}
  \centerline{\includegraphics[width=0.49\textwidth]{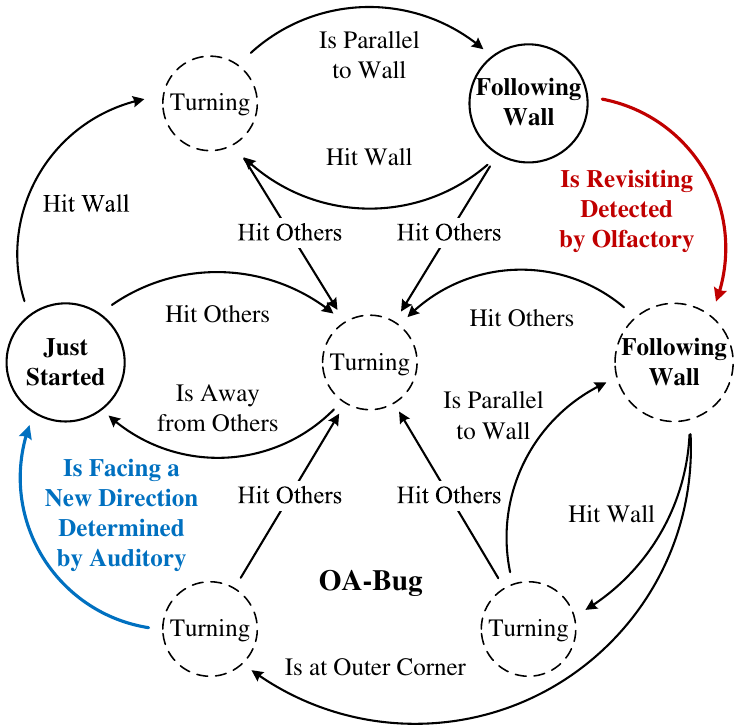}}
  \vspace{-5pt}
  \caption{The Finite State Machine of OA-Bug}
  \label{FSM}
  \vspace{-10pt}
\end{figure}

In order to augment the traditional Bug Algorithm, olfactory and auditory are used to enhance a robot’s sensing ability. When moving, a robot makes auditory signals to indicate its position and leaves a scent to mark the current location as visited. Thus, olfactory enables a robot to know whether the current location $l_{ci} \in S$ has been revisited, and auditory helps to get others’ relative positions for choosing a better direction $\varphi_i$ for turning when it hits a wall or after revisiting a place. Most significantly, each robot is given a preferred direction that can be changed later rather than the direction of a specific target. This is justified as in real search and rescue, exact locations of victims are unknown. Specifically, if robots set off from the center of the site, the preferred direction $\varphi_i$ can be set to $i \times 360\degree/n + \text{bias}$, where $i$ means the $i$th robot in the $n$ robots and the bias can be an arbitrary value, but it is the same for all robots that depart simultaneously. $\boldsymbol{p}_j$ returned from auditory can be converted to $\boldsymbol{c}_j=(\Delta x_j,\Delta y_j)$ where $\rho_j^2 = \Delta x_j^2 + \Delta y_j^2$ and $\tan \theta_j = \frac{\Delta y_j}{\Delta x_j} (\Delta x_j \ne 0)$. Then $\boldsymbol{c}_{i, \text{opp}} = -\sum_j \boldsymbol{c}_j$ is the opposite direction vector of other robots, based on which $\boldsymbol{p}_{i, \text{opp}} = (\rho_{i, \text{opp}}, \theta_{i, \text{opp}})$ is obtained, where $\varphi_i' = \theta_{i, \text{opp}}$ is the new preferred direction.

The finite state machine of OA-Bug is shown in Fig.~\ref{FSM}. The main states are defined as follows.

\textit{1) Just Started:} The robot has just started searching or changed its direction after hitting others or revisiting a place.

\textit{2) Following Wall:} The robot has just hit a wall and is following the wall.

The main triggers and their next states are described below.

\textit{1) Hitting Wall:} If the robot is in the “Just Started” state, it will transfer to the “Following Wall” state. If the robot is in the “Following Wall” state, it will stay in it.

\textit{2) Hitting Others:} If the robot is in the “Following Wall” state, it will transfer to the “Just Started” state. If the robot is in the “Just Started” state, it will stay in it. To reduce the chances of further collisions and revisits, they turn in the same direction (e.g., all turn right) afterward.

\textit{3) At Outer Corner:} If the robot is in the “Following Wall” state and finds that this place has been visited, it will transfer to the “Just Started” state. If the robot is in the “Just Started” state due to revisiting, it will change its orientation.

\textls[-1]{More procedurally, a robot will initialize its state to “Just Started” and will move in the preferred direction $\varphi_i$ until it hits a wall. And then, it will transfer its state to “Following Wall,” after which it moves along the wall. If it encounters the entrance to a room area, it will enter and continue moving along the room. If one room leads to another one, it will search along the wall firstly for the room area inside and then for the room area outside, which is essentially a depth-first search, ensuring that it searches as many room areas as possible, thus increasing the coverage. If it detects a scent when it is at an outer corner of the walls, it means that the area has been visited and the robot is ready to update $\varphi_i$. It will continue following the wall until it arrives at another outer corner of the wall, and then it will update and turn to $\varphi_i'$ and may leave the wall while transferring the state to “Just Started.” $\varphi_i$ is mostly updated via auditory, but if a robot detects that a place has been visited several times (e.g., three times inferred by olfactory), it believes that the nearby area has been thoroughly searched, and should then change to $\varphi_i' + 180\degree$, which is opposite to its preferred direction. So ideally, at this point, the robot will explore an area symmetrical to the center of this area with respect to the departure point, avoiding overlap with the fully-searched area.}

\begin{table*}[t]
  \begin{center}
  \vspace{3pt}
  \caption{\textsc{Chosen Algorithms}}
  \label{chosen algs}
  \begin{tabular}{|c|c|l|l|}
  \hline
  \textbf{Algorithm}                                                     & \textbf{Explanation} & \multicolumn{1}{c|}{\textbf{From Just Started to Following Wall}}                                                                & \multicolumn{1}{c|}{\textbf{In Following Wall and at Outer Corner}}                                                                                                       \\ \hline
  \textbf{Random}                                                        & Blank Control        & only turns at an arbitrary angle                                                                                                 & goes forward or turns at a random angle                                                                                                                                   \\ \hline
  \textbf{Minimum}                                                       & Basically “Bug 0”    & turns in the same direction (e.g., right)                                                                                        & turns to $\varphi_i$                                                                                                                                                      \\ \hline
  \textbf{Auditory Only}                                                   & No Olfactory         & \begin{tabular}[c]{@{}l@{}}updates $\varphi_i'$ and turns left or right\\ according to its direction to follow wall\end{tabular} & updates $\varphi_i'$ and turns toward it                                                                                                                                  \\ \hline
  \textbf{Olfactory Only}                                                     & Advanced “Bug 1”     & turns in the same direction (e.g., right)                                                                                        & \begin{tabular}[c]{@{}l@{}}If a previous place has been visited, it turns to $\varphi_i$\\ (+ 180° if more than 3 times); else it follows wall.\end{tabular}              \\ \hline
  \textbf{\begin{tabular}[c]{@{}c@{}}Olfactory\\+ Auditory\end{tabular}} & Full OA-Bug          & \begin{tabular}[c]{@{}l@{}}updates $\varphi_i'$ and turns left or right\\ according to its direction to follow wall\end{tabular} & \begin{tabular}[c]{@{}l@{}}If a previous place has been visited, it turns to the updated $\varphi_i'$\\ (+ 180° if more than 3 times); else it follows wall.\end{tabular} \\ \hline
  \end{tabular}
  \end{center}
  \vspace{-20pt}
\end{table*}

\section{Software Simulation}

To measure the performance of OA-Bug, we first performed a software simulation. 

\subsection{Simulation Environment}

To acquire complete control over the entire simulation process, we built the environment nearly from scratch\footnote{\textls[-8]{\texttt{\url{https://github.com/kevintsq/SwarmSearchSimulation}}}.}, which consists of the following main parts.

\textit{1) Site Generator:} The Site Generator is implemented to generate a reasonable site randomly. “Reasonable” means that the generated site must conform to the definition given in the above section. “Randomly” means that the size, layout, relationship between rooms, and the width of corridors are almost different for each generated site, avoiding the suspicion of elaborately designing a site for better results. The width and height of the site and the number of rooms and victims are parameters that can be passed to the generator for each generation, enabling us to perform large-scale tests efficiently and analyze the relationship between different configurations of the site and the statistical results.

\textit{2) Layout Module:} The Layout Module creates objects according to the site generated by the generator or by hand. If graphics mode is enabled, it also draws the layout on the screen, as shown in Fig.~\ref{header} and Fig.~\ref{defs fig}.

\textls[-1]{\textit{3) Robot Module:} The Robot Module contains different robot objects implemented in different algorithms. If graphics mode is enabled, robots will be drawn on the screen. Note that although the robot must know its absolute position so it can be drawn on the screen, in order to obey our scenario setup, it never builds a map or uses its position to decide its next action.}

\subsection{Simulation Design}
The purpose of the simulation is to compare the performances of different algorithms and analyze the relationship between different simulation configurations and the statistical results. The chosen algorithms are briefly described in Table~\ref{chosen algs}. The selected configurations are listed in Table~\ref{site conf}.

\begin{table}[t]
  \vspace{5pt}
  \caption{\textsc{Site Configurations}}
  \vspace{-20pt}
  \label{site conf}
  \begin{center}
    \begin{tabular}{|c|c|c|c|c|c|}
      \hline
      \textbf{\begin{tabular}[c]{@{}c@{}}Configur-\\ation\end{tabular}} & \textbf{\begin{tabular}[c]{@{}c@{}}Width*\\ ($\times \frac{4}{15}$m)\end{tabular}} & \textbf{\begin{tabular}[c]{@{}c@{}}Height\\ ($\times \frac{4}{15}$m)\end{tabular}} & \textbf{\begin{tabular}[c]{@{}c@{}}Area\\ (m$^2$)\end{tabular}} & \textbf{\begin{tabular}[c]{@{}c@{}}\# of\\ Rooms\end{tabular}} & \textbf{\begin{tabular}[c]{@{}c@{}}Max Actions\\Per Robot*\end{tabular}} \\ \hline
      \textbf{Large}                                                    & 120                                                          & 60                                                            & 512                                                          & 120                                                            & 4000                                                                              \\ \hline
      \textbf{Medium}                                                   & 80                                                           & 40                                                            & 228                                                          & 60                                                             & 2000                                                                              \\ \hline
      \textbf{Small}                                                    & 60                                                           & 30                                                            & 128                                                          & 30                                                             & 1000                                                                              \\ \hline
      \end{tabular}
  \end{center}
  *Units and values are chosen based on the size of real-world sites and the aesthetics of graphic display.
  \vspace{-10pt}
\end{table}

\textls[-5]{In the simulation, for each configuration, the site generator was used to generate 100 different sites. For each site, we applied the above five algorithms using 2, 4, 6, 8, and 10 robots departing from the center and the boundary of the site. For each run, the total coverage, coverage per robot, collision count, and other necessary data were recorded per 100 actions. If a robot has entered a room, then it is considered to have covered a room. A round of simulation terminates when all rooms are covered, or each robot has reached its maximum number of actions.}

\begin{figure*}[t]
  \vspace{1pt}
  \centering
  \begin{subfigure}{\textwidth}
    \centering
    \includegraphics[width=0.9\textwidth]{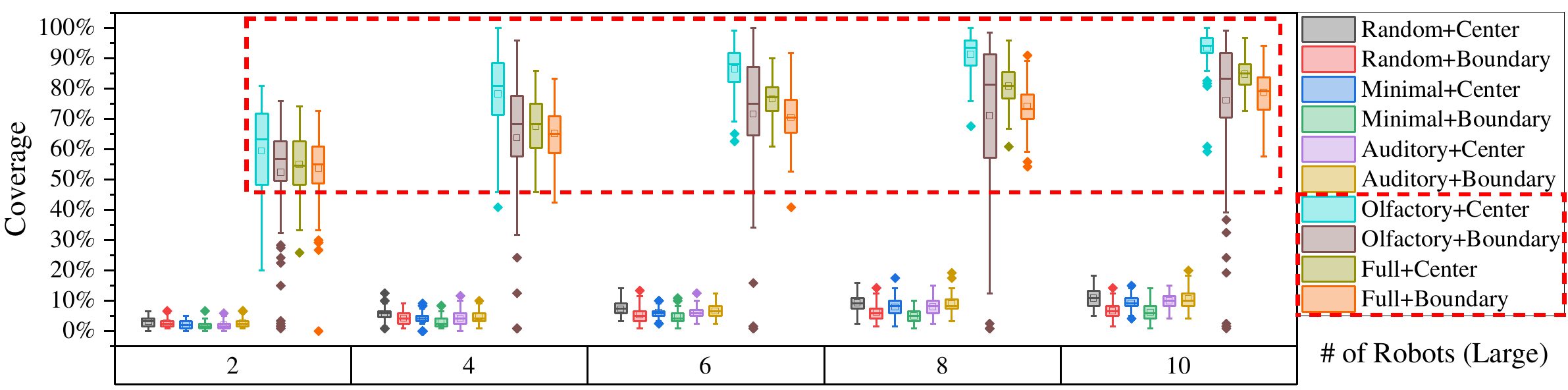}
  \end{subfigure}
  \begin{subfigure}{\textwidth}
    \centering
    \includegraphics[width=0.9\textwidth]{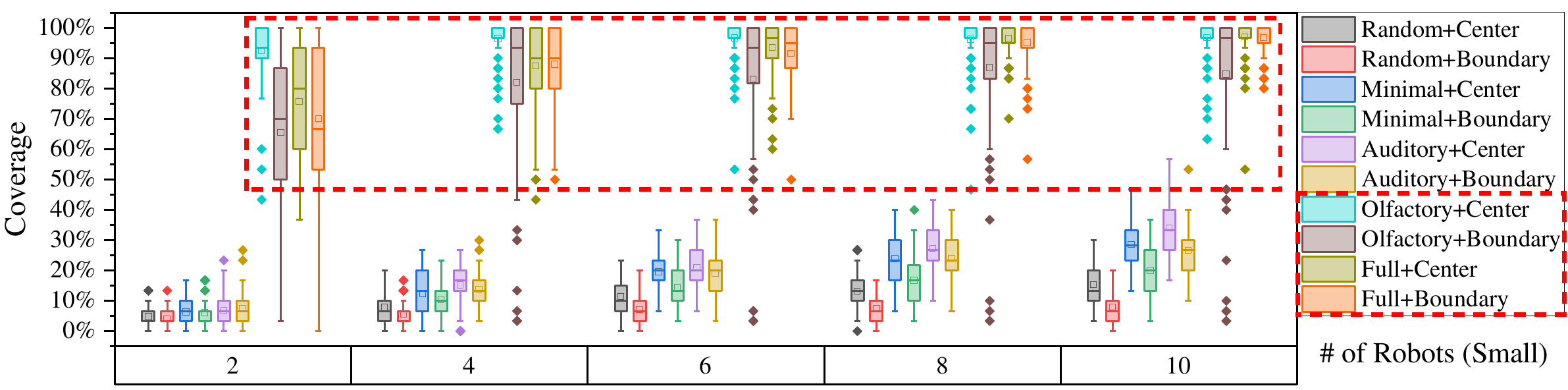}
  \end{subfigure}
  \begin{subfigure}{0.325\textwidth}
    \centering
    \includegraphics[width=\textwidth]{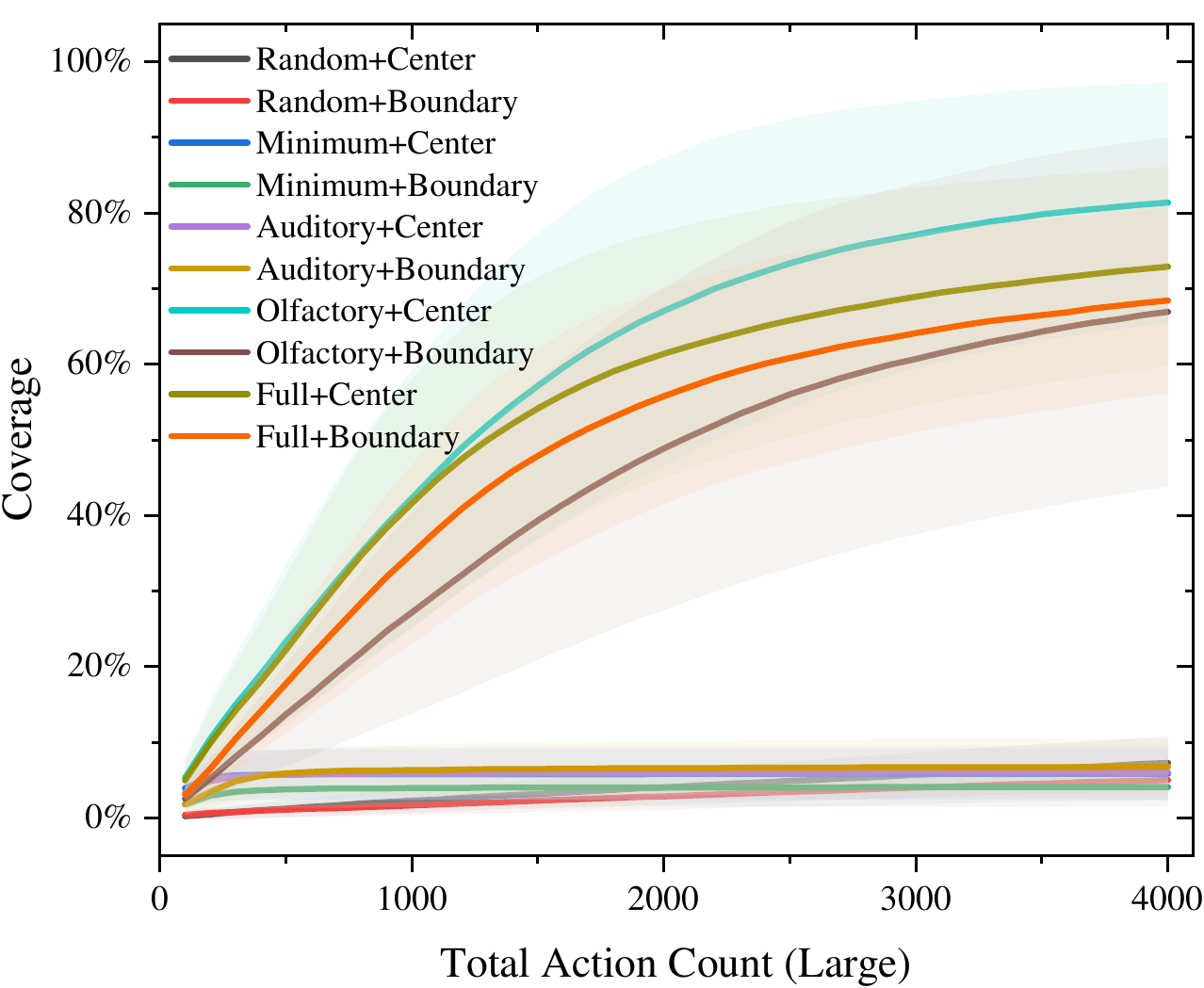}
  \end{subfigure}
  \begin{subfigure}{0.325\textwidth}
    \centering
    \includegraphics[width=\textwidth]{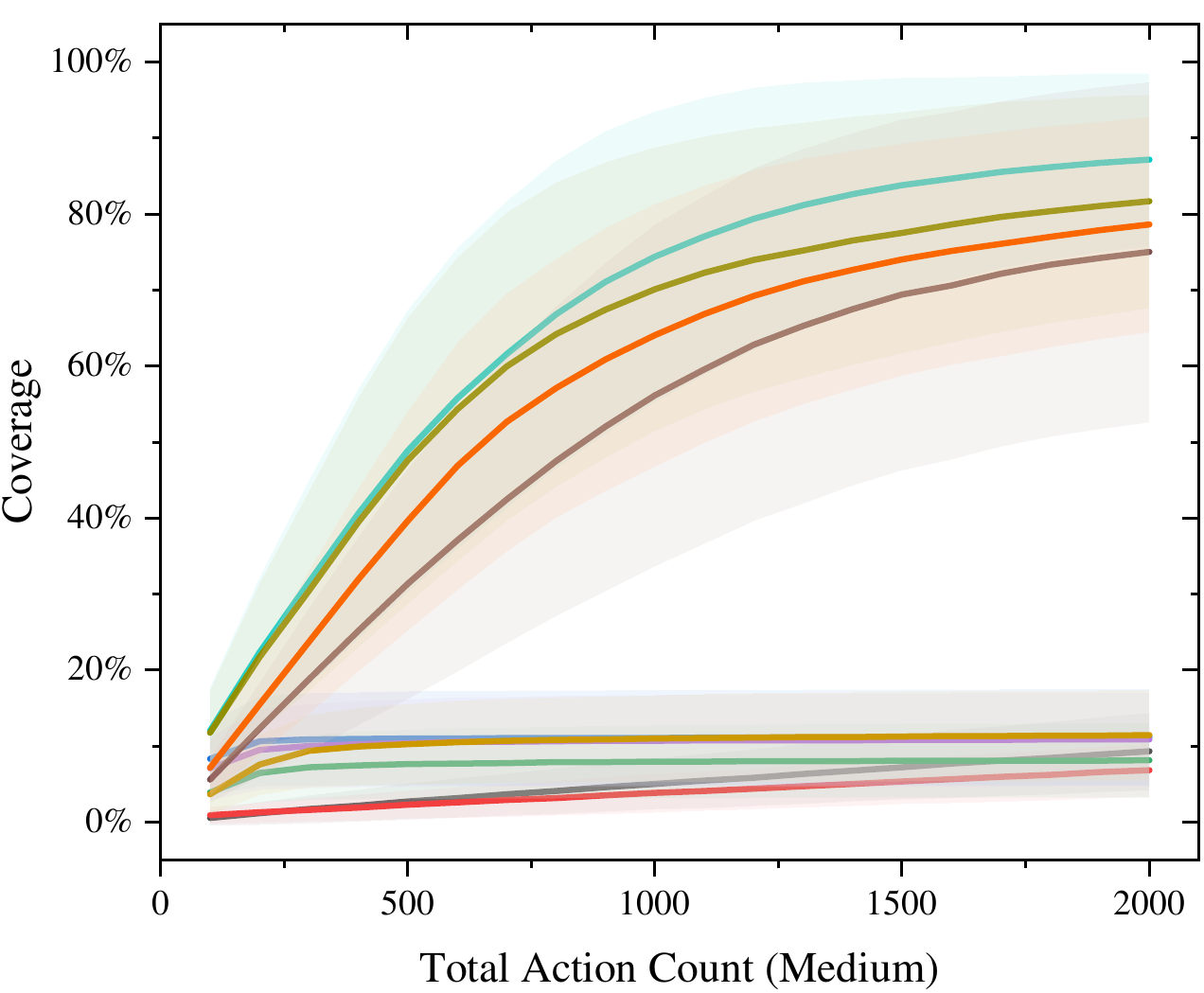}
  \end{subfigure}
  \begin{subfigure}{0.325\textwidth}
    \centering
    \includegraphics[width=\textwidth]{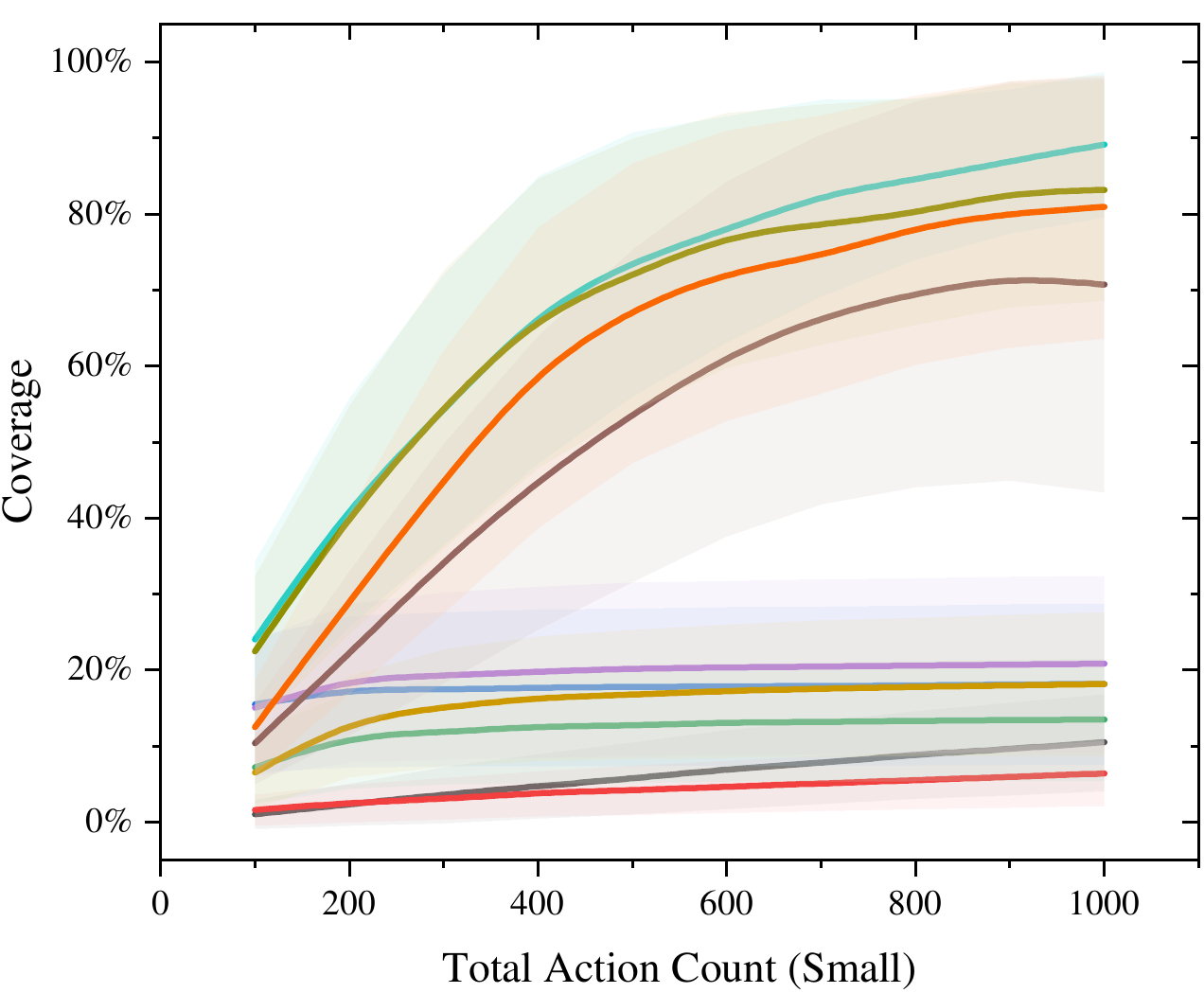}
  \end{subfigure}
  \vspace{-5pt}
  \caption{Simulation Results of Searching \textit{(Note that the average coverage associated with total action count is relatively lower than the coverage shown elsewhere, because if a run has achieved 100\% coverage before the maximum action limit is reached, the simulation and recording ends early.)}}
  \label{results}
\end{figure*}

\begin{figure*}[t]
  \centering
  \begin{subfigure}{0.325\textwidth}
    \centering
    \includegraphics[width=\textwidth]{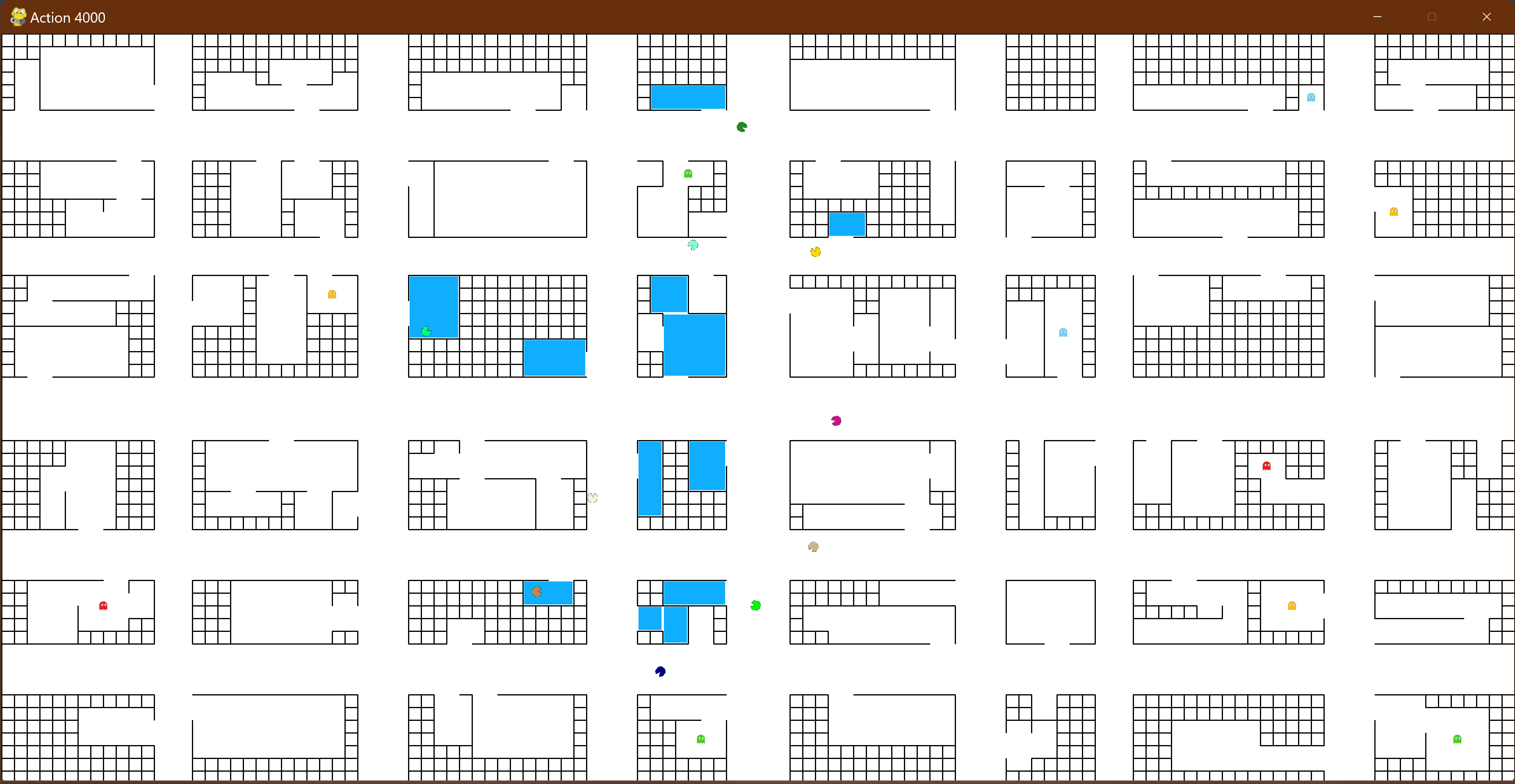}
    \caption{Random (Blank Control)}
  \end{subfigure}
  \begin{subfigure}{0.325\textwidth}
    \centering
    \includegraphics[width=\textwidth]{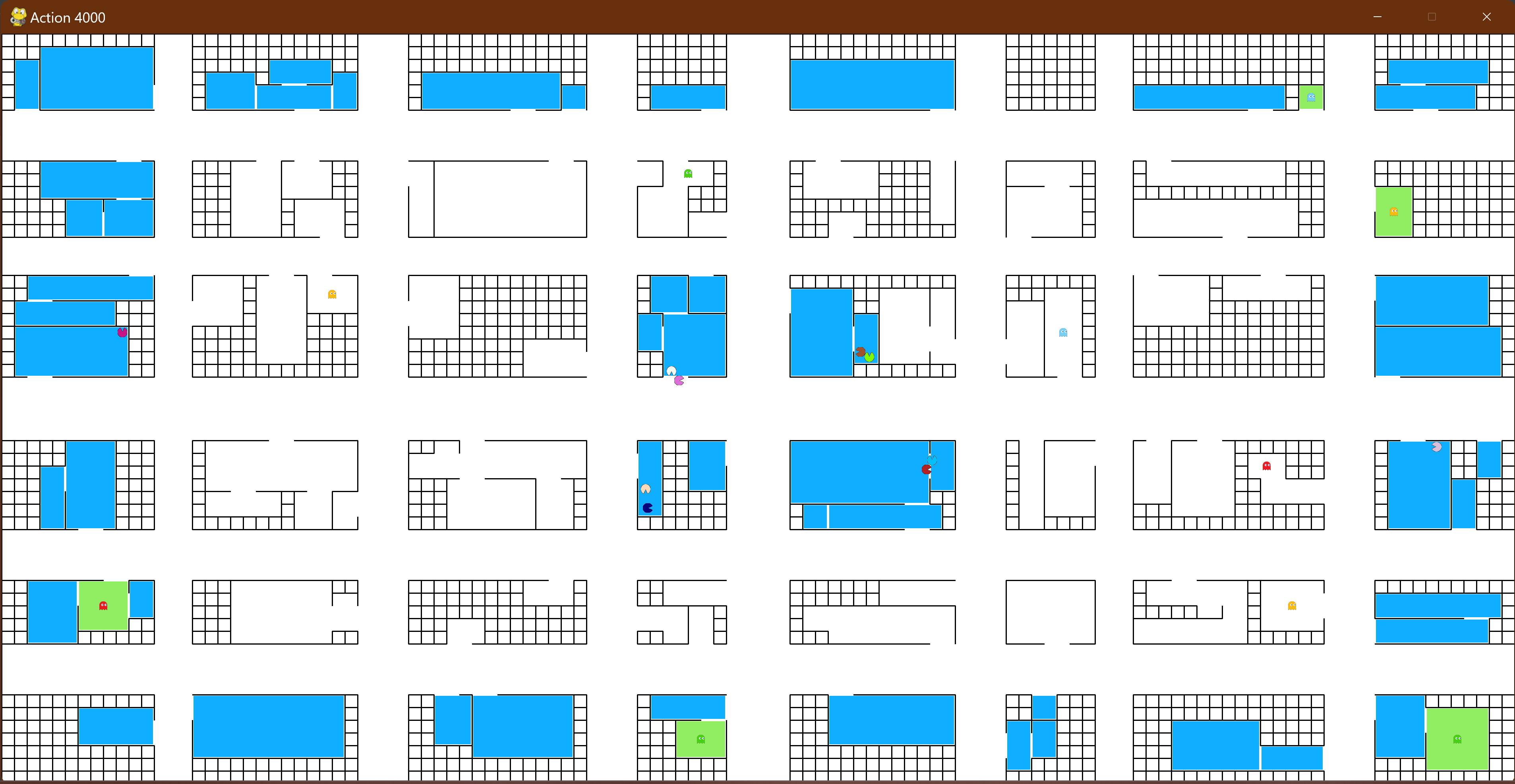}
    \caption{Schematic Representation of SGBA's Coverage*}
  \end{subfigure}
  \begin{subfigure}{0.325\textwidth}
    \centering
    \includegraphics[width=\textwidth]{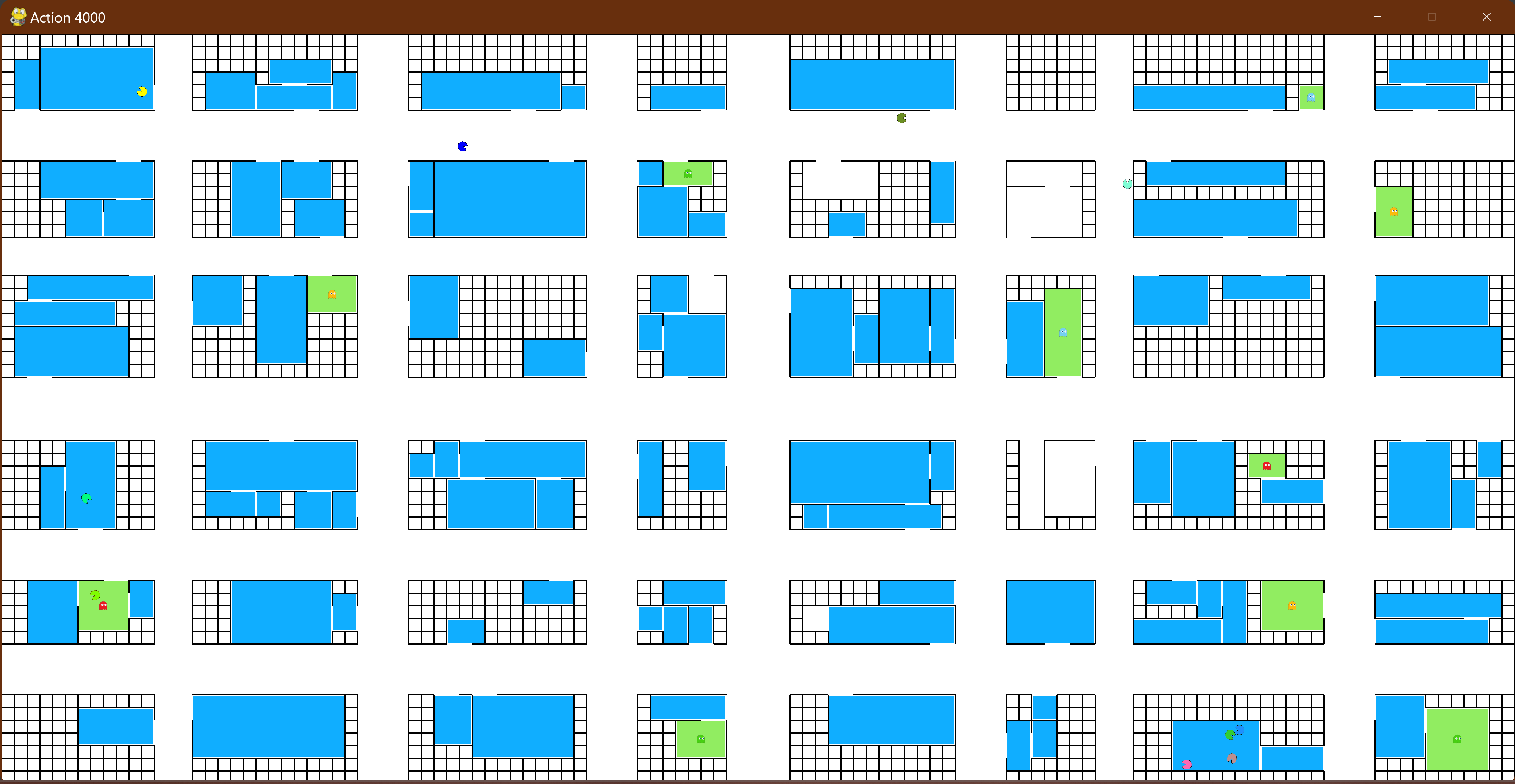}
    \caption{One of the Best Results of OA-Bug}
  \end{subfigure}
  \vspace{-5pt}
  \caption{Sample Result Screenshots of Different Algorithms \textit{(*Not yet fully ported to our setting. Just a general illustration based on FSM in~\cite{c1}.)}}
  \label{comparasion}
  \vspace{-10pt}
\end{figure*}

\subsection{Simulation Results and Discussions}

We performed simulations on different machines and got almost identical results, which means the results are reproducible. The results are shown in Fig.~\ref{results}. We mainly use coverage to evaluate our algorithm, as improving it is our primary goal to develop the algorithm. Victims will not be discussed because no special mechanisms are designed for them, but the interface is reserved for future works. In short, the results are positive evidence that our algorithm is effective and can achieve a high coverage rate.

First, the coverage of “Olfactory Only” and OA-Bug are much higher than others. Especially in Large Configuration, the coverage of “Minimum” and only “Auditory Only” are almost as low as “Random,” which means being able to know where has been visited and change direction accordingly has a tremendous contribution to the coverage.

In addition, the more robots are used, the higher the coverage and the lower the standard deviation of coverage is, but the increasing rate of coverage relative to the number of robots decreases when the number of robots increases, which is especially evident in small and medium configurations, suggesting the number of robots should be carefully chosen according to the size of the site in real-life scenarios.

Besides, the coverage when robots depart from the center is almost higher than that when robots depart from the boundary, and the number of actions required to reach the same coverage when robots depart from the center is less than that when robots depart from the boundary, which is obvious because robots can spread more evenly when they depart from the center. But in realistic scenarios, a site’s entrance is more likely to be at the boundary. So it is essential to have high coverage when departing from the boundary.

Furthermore, the coverage of “Olfactory Only” is higher than that of OA-Bug when departing from the center but lower when departing from the boundary. The coverage of OA-Bug when departing from the boundary is much higher than that of “Olfactory Only” when departing from the boundary in Small Configuration and is almost equal to that of OA-Bug when departing from the center. The number of actions required to reach the same coverage when OA-Bug departs from the boundary is less than that when “Olfactory Only” departs from the boundary. The utilization of auditory accounts for these results. When first hitting the wall, if robots depart from the center, since there is enough space for them to spread, it is more important for robots to turn in the same direction (e.g., all turn right). This allows them to search the site in a clockwise or counterclockwise manner, thus reducing collisions. “Olfactory Only” is implemented this way, so the coverage, when they depart from the center, is higher. If robots depart from the boundary, since there is not enough space for them to spread out, it is more important for robots to turn in different directions so they can spread out more evenly to reduce revisits. OA-Bug is implemented this way, so the coverage, when they depart from the boundary, is higher. In practice, such basic algorithms can be integrated or tuned to achieve high coverage in all scenarios, and most observations still hold.

Despite the assistance of the Wi-Fi home beacon and the departure of robots from the center in SGBA~\cite{c1}, our algorithm has higher coverage. As their site is smaller and less complex, their results would be worse if ported to our platform, although an in-depth comparison to SGBA~\cite{c1} is yet to be reported. Fig.~\ref{comparasion} shows a sample screenshot of the results of the different algorithms.

\section{Real-World Experimental Results}

To demonstrate that OA-Bug works in the real world and verify that the results are consistent with the simulated ones, we implemented the entire system on hardware and conducted experiments\footnote{\texttt{\url{https://github.com/kevintsq/OA-Bug-Robot}}.}.

\subsection{Hardware Setup and Testing}
Four robots are involved in the experiment. Each robot (Fig. 1) is equipped with a Raspberry Pi 4B as the central controller, an MP503 Ethanol Sensor\footnote{\texttt{\url{https://wiki.seeedstudio.com/Grove-Air_Quality_Sensor_v1.3/}}, last visited on 2023-7-18.} as its olfactory, a Silicon Labs BG22-RB4191A Bluetooth Antenna Array Radio Board\footnote{\texttt{\url{https://www.silabs.com/development-tools/wireless/bluetooth/bg22-rb4191a-bg22-bluetooth-dual-polarized-antenna-array-radio-board}}, last visited on 2023-7-18.} as its auditory locator, a Silicon Labs SLTB010A EFR32BG22 Thunderboard Kit\footnote{\texttt{\url{https://www.silabs.com/development-tools/thunderboard/thunderboard-bg22-kit}}, last visited on 2023-7-18.} as its sound tag, a Slamtec Rplidar A1 for wall following and obstacle detection, an MPU6050 gyro for direction control, a 3D-printed ethanol tank and a 3-watt pump for ethanol release.

With a unique MAC address, each sound tag emits signals using Bluetooth 5.1 protocol and can be received by each auditory locator. Afterward, the signal data is processed using the angle of arrival (AoA) algorithm in the Gecko library\footnote{\texttt{\url{https://github.com/kevintsq/BluetoothAoaLocator}}.}, and the tag’s direction relative to the locator can be precisely obtained with less than 5° error\footnote{Officially claimed and demonstrated by Silicon Labs on \texttt{\url{https://youtu.be/c3XqbEKmNcM}}, last visited on 2023-7-18.} in real-time. Each robot ignores signals from its own tag. Although the Bluetooth signal is used to mimic the audible sound, no state data is shared. Our implementation takes advantage of the new Bluetooth 5.1 Direction Finding standard that can provide us with sub-meter precision, a much longer range (approximately 30–50 meters range with 4–0 walls blocking respectively tested by us), and more robust source localization and tracking performance than real sound~\cite{c30}.

\textls[-1]{The position of the gas sensor and ethanol is elaborately managed so that a robot’s ethanol cannot interfere with its own sensor. The pump releases 75\% ethanol every 5 seconds for 0.5 seconds. The analog output of the gas sensor is sampled by a 10-bit ADC, ranging from 0 to 1023. Typically, the digital output is below 100, and if the sensor passes near (within one meter of) ethanol released in the manner described above, the output will typically exceed 100 within 5 seconds and then drop below 200 within 10 seconds. Thus, we set the threshold to 200 for determining whether the site has been reached. Note that the experiment is carried out in a site with closed windows and mild AC controlling the temperature around 27 ℃, so there is barely any air turbulence. These parameters may vary if the experiment is conducted on a different site.}

Although equipped with LiDAR, robots do not build maps. For wall following, raw scans are first filtered to remove outlier points, after which line segments are extracted using a split-and-merge procedure and refined via weighted least-squares line fitting to estimate line parameters and their associated covariance matrices~\cite{c32}. For obstacle avoidance, LiDAR returns are geometrically clustered to detect obstacles and trigger local avoidance behaviors~\cite{obstacle}.

\subsection{Experimental Design}
To verify the effectiveness of our algorithm, we choose the site with at least one “isolated island” (see Sec.~\ref{defs} for definition) so that olfactory and auditory must be used to achieve high coverages. The site is the most complicated one available to us, although robots cannot enter rooms due to doors closure. This is justifiable because the rooms are densely packed with furniture that may interfere with the basic wall-following behavior, which is not the core of our experiment; the focus is to test the olfactory and auditory. After entering a room, a robot can always leave as long as it follows the wall. Thus, if a robot has followed a wall of a room, then it is considered to have covered a room.

The robots set off from the boundary of the site in their initial directions and then search the site under surveillance. We stop a round of experiments if the site has been thoroughly explored or 20 minutes have passed.

\begin{figure}[t]
  \vspace{3pt}
  \centerline{\includegraphics[width=0.49\textwidth]{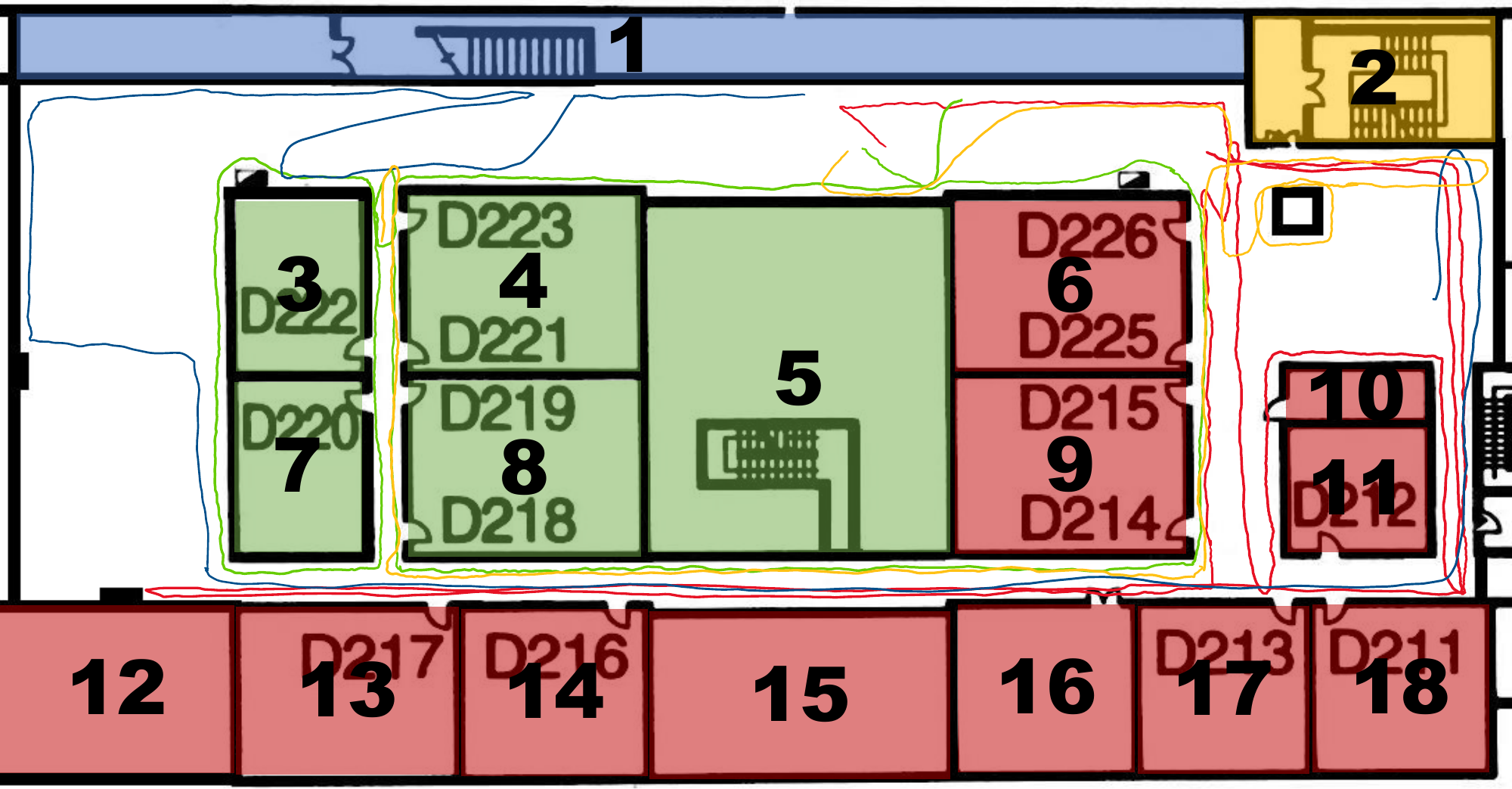}}
  \vspace{-5pt}
  \caption{The Best Result in the Real-World Experiment}
  \footnotesize \textit{(Rooms are numbered and paths are colored to correspond with each robot. \textls[-1]{The color of a room is the same as the color of the robot that visited it first.})}
  \label{paths}
  \vspace{-10pt}
\end{figure}

\subsection{Experimental Results and Discussions}
\textls[-1]{Among the six rounds of valid experiments, the robots covered an average of 15.167 rooms out of the 18 rooms, with a coverage of 84.26\%. The coverage is also higher than that of SGBA~\cite{c1}, despite their real-world site not having any “isolated islands.” The result shows that our implementation works effectively in real-world scenarios. With the help of olfactory and auditory, the robots will neither be stuck in narrow corridors nor visit a place that has been visited repeatedly. The best result for a round of experiments is shown in Fig.~\ref{paths}.}

The reasons the coverage is lower than that in simulations are as follows. First, real-world walls are not always flat and straight, rendering some walls recognized as obstacles. However, the coverage is still acceptable even if our implementation of wall following and obstacle avoidance does not involve expensive approaches such as machine learning. Second, some robots experienced hardware failure during some experiments. Nevertheless, those well-functioned robots contributed much to the coverage, proving the algorithm’s robustness to some robots’ hardware failure.

\section{Conclusion and Future Work}

We propose the OA-Bug for a swarm of autonomous robots with olfactory and auditory to explore a denied environment efficiently where GNSS, mapping, data sharing, and central processing are unavailable. The simulation and the experiment proved that the OA-Bug is effective and efficient. In the future, we can apply more tasks to OA-Bug than just searching, such as returning to the departure point after a search is completed, a robot is low on power, or they are being summoned. Furthermore, we may analyze and optimize the algorithm by building models from the macro perspective rather than the micro behavior level, develop more precise sensor models, and port the system to UAVs.



\section*{Acknowledgment}

We thank Rongkun Jiang, Sijie Wang, Yue Li, and Lichao Wang for their help with the experiments.

\bibliographystyle{IEEEtran}
\bibliography{main}

\end{document}